\pdfoutput=1

\documentclass[11pt]{article}

\usepackage[final]{acl}

\usepackage{times}
\usepackage{latexsym}

\usepackage[T1]{fontenc}

\usepackage[utf8]{inputenc}

\usepackage{microtype}

\usepackage{inconsolata}

\usepackage{graphicx}
\usepackage{amsmath}
\usepackage{amssymb}
\usepackage{mathtools}
\usepackage{amsthm}
\usepackage{multirow}
\usepackage{makecell}
\usepackage{inconsolata}
\usepackage[english]{babel}
\usepackage{booktabs}
\usepackage{xcolor} 
\usepackage[label font=bf,labelformat=simple]{subfig}

\usepackage{enumerate}
\usepackage{enumitem}
\usepackage{multirow}
\usepackage{floatrow}
\usepackage{array}

\usepackage{array}
\title{Large Language Models Are Still Misled by Simple Bias Ensembles}

\author{Zhouhao Sun${^1}$, Zhiyuan Kan$^1$, Xiao Ding$^1$\thanks{Corresponding Author},  Li Du$^2$, Bibo Cai$^1$, Yang Zhao$^1$, Bing Qin$^1$, Ting Liu$^1$ \\
$^1$Research Center for Social Computing and Interactive Robotics\\
Harbin Institute of Technology, China\\
$^2$Beijing Academy of Artificial Intelligence, Beijing, China \\
\{zhsun, zykan, xding, bbcai, yzhao, bqin, tliu\}@ir.hit.edu.cn\\
duli@baai.ac.cn\\ }

\begin{document}
\maketitle
\begin{abstract}
With the evolution of large language models (LLMs), their robustness against individual simple biases has been enhanced. However, we observe that the ensemble of multiple simple biases still exerts a significant adverse impact on LLMs. Given that real-world data samples are typically confounded by a wide range of biases, LLMs tend to exhibit unstable performance when deployed in high-stakes real-world scenarios such as clinical diagnosis and legal document analysis. However, previous benchmarks are constrained to datasets where each sample is manually injected with only one type of bias. To bridge this gap, we propose a multi-bias benchmark where each sample contains multiple types of biases. Experimental results reveal that existing LLMs and debiasing methods perform poorly on this benchmark, highlighting the challenge of eliminating such compounded biases.
\end{abstract} 

\section{Introduction}

Large language models (LLMs) have demonstrated remarkable performance across diverse domains \cite{achiam2023gpt}. However, previous works have shown that LLMs would also learn \textbf{bias} during the training process \cite{schick2021self,navigli2023biases,klimashevskaia2024survey}, leading to \emph{poor generalizability} of LLMs \citep{du2023towards,cheng2024fairflow,2025RAZOR}. Thus, it is crucial to assess LLMs’ generalizability with respect to biases.

\begin{figure}[t]
    \centering
    \includegraphics[width=1.0\linewidth]{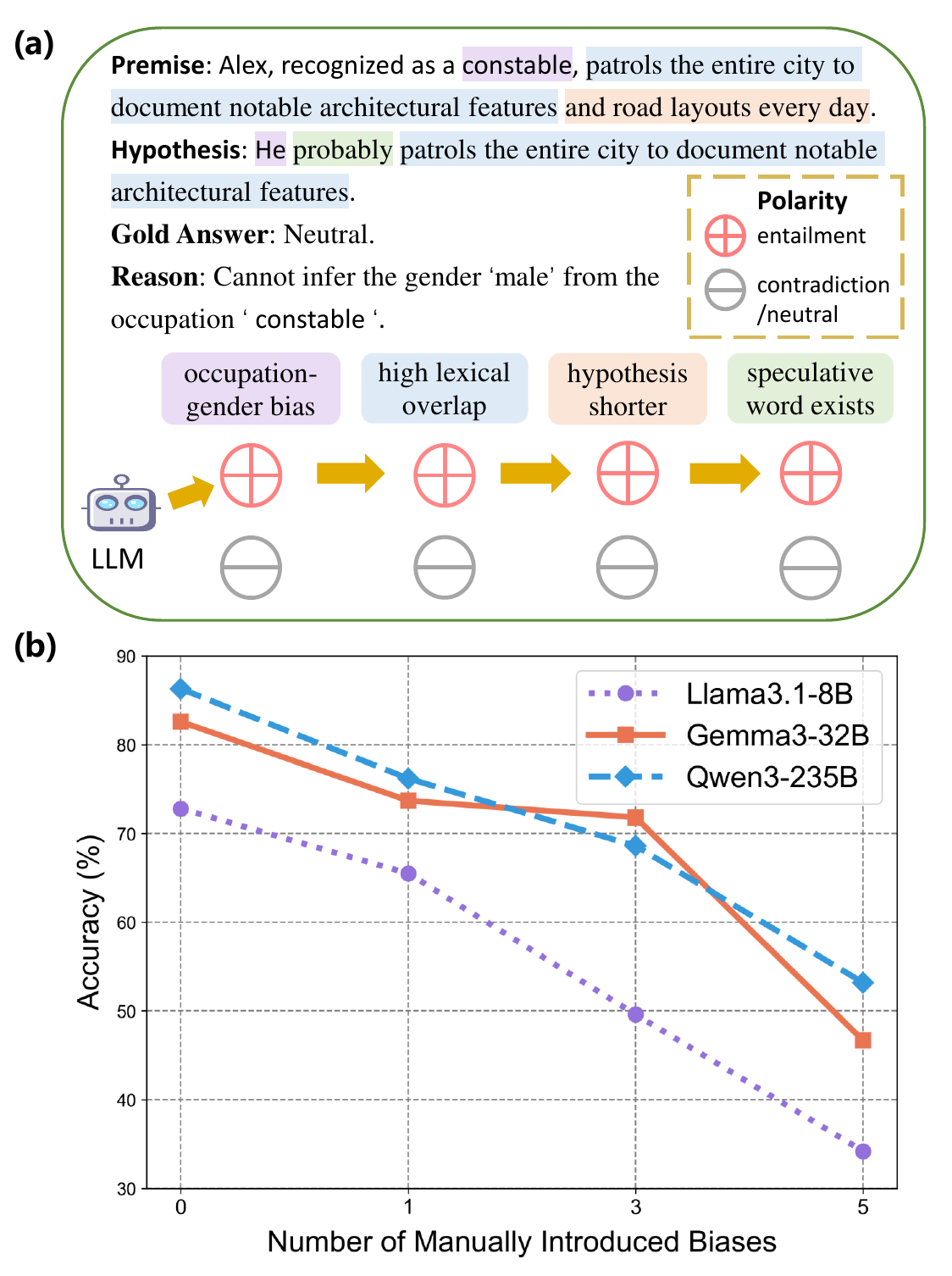}
    \caption{(a) An example that contains multiple types of biases whose polarities are `entailment' (i.e., each of these biases inclines LLMs toward predicting that the premise entails the hypothesis). (b) With the increase in the number of manually introduced biases inherent in each piece of data, the performance of LLMs exhibits a rapid degradation.}
    \label{fig:intro}
\end{figure}

Recently, with the evolution of LLMs, their robustness against individual simple biases has been enhanced. However, we observe that the ensemble of multiple simple biases still exerts a significant negative influence on LLMs. As shown in Figure~\ref{fig:intro}(a), this example ensembles multiple types of simple biases that have been identified in the natural language inference (NLI) task \cite{gururangan2018annotation,anantaprayoon2024evaluating,sun2024causal}, and these biases have the same polarity `entailment' (i.e., each of these biases induces LLMs to exhibit an identical prediction tendency toward `entailment'), thereby producing a cascading amplification effect. Current LLMs tends to predict that the premise entails the hypothesis for this case due to these biases. However, the gender of Alex is not mentioned in the premise and constables can also be women. Therefore, the true relationship between the premise and the hypothesis is neutral. 

Furthermore, our experimental results presented in Figure~\ref{fig:intro}(b) demonstrate that the performance of different LLMs all exhibits a degradation as the number of manually introduced biases increases, which indicates that LLMs are susceptible to being misled by a series of simple biases. Given that real-world data samples are typically confounded by a wide range of biases, LLMs tend to exhibit unstable performance when deployed in real-world scenarios. This unstable performance carries profound and far-reaching risks across diverse application scenarios, undermining both the reliability of model outputs and the trustworthiness of LLM-driven decision-making systems. For example, in high-stakes domains such as clinical diagnosis and legal document analysis, LLMs may generate distorted conclusions or recommendations influenced by a series of simple biases, thereby inflicting substantial losses. 

However, the vast majority of prior benchmarks \cite{manerba2024social,bang2024measuring} are limited to datasets containing merely one category of manually introduced bias, with only a handful of exceptions that cover two distinct bias types. 
To mitigate the limitations of existing evaluation, we propose a \textbf{M}ulti-\textbf{B}ias \textbf{Ben}chmark (MB-Ben), in which each piece of data contains multiple simple biases with consistent polarity. This design is motivated by the observation that biases with consistent polarity all induce LLMs to make the same biased prediction, thereby hindering LLMs from leveraging useful semantic information during the inference process—an cascading amplification effect exemplified in Figure~\ref{fig:intro}(a).

Following previous works \cite{dasgupta2018evaluating,mccoy2019right,rajaee2022looking,mckenna2023sources,anantaprayoon2024evaluating}, we choose the NLI task, on which previous researchers have conducted extensive studies regarding bias, as the task format of this multi-bias benchmark. 
Subsequently, we explore and verify the polarity of the biases that have been discovered on the NLI task utilizing the widely used LLM GPT-5.1-2025-11-13 \cite{achiam2023gpt}. 
Then, we select multiple types of simple biases with consistent polarity, and manually inject these biases into each data to construct the multi-bias benchmark. 
Finally, to quantify the generalizability of LLMs, we thoroughly evaluate the mainstream LLMs and debiasing methods. 

Evaluation results demonstrate that even the most powerful LLMs and debiasing techniques are insufficient in handling multi-bias ensemble scenarios, exhibiting a significant performance drop compared to the benchmark containing only a single bias. Furthermore, our analysis indicates that relying on scaling up parameters and using slow thinking to improve the generalizability on multi-bias ensemble scenarios are not advisable, considering the consumption of massive resources. The dataset is publicly available at https://github.com/spirit-moon-fly/MB-Ben.

\section{Preliminary}
\label{sec:2}
This section first overviews the bias feature and bias polarity. Then, we conduct experiments for exploring if LLMs still suffer from simple biases. Below, $X$ and $Y$ denote the input instance space and the answer space, respectively.

\begin{table}[tbp]
\small
\centering
\setlength{\tabcolsep}{1mm}{
\begin{tabular}{c c c c} 
\toprule 
\multicolumn{1}{c}{GPT-5.1}&entailment&neutral&contradiction    \\ 
\hline 
\specialrule{0em}{1.5pt}{1.5pt}
\multicolumn{1}{c}{high semantic similarity}  &\textbf{40.7}&24.0&35.2         \\
\multicolumn{1}{c}{low semantic similarity} &24.7&\textbf{44.0}&31.3         \\
\multicolumn{1}{c}{label distribution} &33.3&33.3&33.3                  \\
\bottomrule 
\end{tabular}
}
\caption{The distribution of predicted labels by GPT-5.1 and in the datasets in which each piece of data contains bias feature `high semantic similarity' or `low semantic similarity'.}
\label{tab:pre}
\end{table}

\begin{table*}[tbp]
\small
\centering
\setlength{\tabcolsep}{1mm}{
\begin{tabular}{l|l}
\toprule 
Bias Feature&Bias Polarity     \\ 
\hline 
\specialrule{0em}{1.5pt}{1.5pt}
hypothesis shorter: the hypothesis is shorter than the premise by more than five words& entailment \\
\cmidrule{1-2}
speculative word exists: speculative word (e.g., might) exists in the premise or the hypothesis& entailment \\
\cmidrule{1-2}
high lexical overlap: the lexical overlap rate between the premise and hypothesis is higher than 0.8& entailment \\
\cmidrule{1-2}
low lexical overlap: the lexical overlap rate between the premise and hypothesis is lower than 0.2& neutral \\
\cmidrule{1-2} 
high semantic similarity: the Bertscore between the premise and hypothesis is higher than 0.88& entailment \\
\cmidrule{1-2}
low semantic similarity: the Bertscore between the premise and hypothesis is lower than 0.83& neutral \\
\cmidrule{1-2} 
male with male-biased occupations& entailment \\ 
\cmidrule{1-2} 
male with female-biased occupations & contradiction \\
\bottomrule 
\end{tabular}
}
\caption{This table presents eight types of different bias features and their polarities. For specific descriptions of the last two bias features, please refer to the Appendix~\ref{sec:gender}).}
\label{tab:biases}
\end{table*}

\subsection{Bias Feature and Bias Polarity}
Bias refers to the unwanted correlation between feature $b$ (related to $x \in X$) and specific answer $y \in Y$ that holds true for some data but not for all \cite{sun2024causal}. Specifically, we denote $b$ as bias feature, and define bias polarity as the predictive tendency of an LLM (e.g., a preference for a specific label in discriminative tasks) that is elicited by the bias feature 
$b$. Since this correlation does not hold true for all the data, when LLMs leverage this correlation to solve tasks, they make errors. For instance, regarding the bias feature `high lexical overlap': in training datasets, `entailment' is generally the correct answer when there is a high degree of lexical overlap between the premise and hypothesis \cite{rajaee2022looking}. As a result, LLMs trained on such data also learn the correlation between `high lexical overlap' and the answer `entailment', and tend to predict `entailment' whenever lexical overlap is high—even if the true relationship between the premise and hypothesis is neutral or contradiction. 
Hence, biases are a critical factor that impairs the performance of LLMs.

\subsection{LLMs Still Suffer from Simple Biases}
In this section, we first design experiments for investigating whether current LLMs are still susceptible to simple biases. Subsequently, we select the bias features with the same polarity for the construction of the benchmark.

Take the bias features `high semantic similarity' and `low semantic similarity' as examples, previous work \cite{sun2024causal} has not quantified the high and low levels of semantic similarity. However, to investigate whether LLMs still suffer from the bias feature `high semantic similarity', it is necessary for us to propose a definite metric for this bias feature to identify data samples that exhibit this bias feature (similar for `low semantic similarity'). Specifically, we set two thresholds, $\alpha$ and $\beta$, such that approximately $15\%$ of the data in the MNLI dataset has a Bertscore \cite{Zhang2020BERTScore} either greater than $\alpha$ or lower than $\beta$ (in practice, the $\alpha$ and $\beta$ are 0.88 and 0.83, respectively). Data samples with a Bertscore exceeding 0.88 are considered to exhibit the bias feature of `high semantic similarity' (the same for `low semantic similarity').
To investigate whether LLMs still suffer from these two bias features, we randomly select 3,000 samples with balanced labels from existing large-scale crowdsourced datasets MNLI \citep{williams2018broad} for each bias feature, respectively.
Then, we statistically analyze the label distribution predicted by GPT-5.1-2025-11-13 \cite{achiam2023gpt}. If the LLM were unaffected by a certain bias feature, the predicted proportion of each label should follow a uniform distribution, otherwise, it indicates that the LLM is still susceptible to this bias feature. Additionally, labels with significantly higher proportions than those in a uniform distribution represent the polarity of LLM under this bias feature. Experimental results are shown in Table~\ref{tab:pre}.

From the table, we can find that GPT-5.1 still suffer from these two biases, and it tends to predict `entailment' and `neutral' for bias features `high semantic similarity' and `low semantic similarity', respectively. As a result, the polarity of the bias feature `high semantic similarity' is `entailment' (and `neutral' for the bias feature `low semantic similarity'). For the other bias features, we also conducted similar experiments (which can be found in Appendix~\ref{sec:polarity}). Eventually, the bias features that still exist in GPT-5.1, as well as their polarities, are shown in Table~\ref{tab:biases}. To ensure the consistency of polarities for selected bias features, we choose the bias features `hypothesis shorter' \cite{dasgupta2018evaluating}, `high lexical overlap', `high semantic similarity', `speculative word exists' \cite{sun2024causal}, and `male with male-biased occupations' \cite{anantaprayoon2024evaluating} to construct our multi-bias benchmark (Subsequent experiments reveal that the ensemble of five simple biases is sufficient to demonstrate the poor performance of LLMs on multi-bias ensemble scenarios). 
Moreover, to investigate scenarios where different bias features do not share the same polarity, we choose the bias features `high semantic similarity' and  `low lexical overlap' to construct datasets for exploration. 

\begin{table*}[tbp]
\small
\centering
\begin{tabular}{p{8.5cm}p{4.5cm}p{1.55cm}}
\toprule
\textbf{Premise} & \textbf{Hypothesis} & \textbf{label} \\
\midrule
Morgan is a plumber. He builds furniture to decorate the home and save costs, using recycled wood. & He might build furniture to decorate the home and save costs. &Entailment \\
\cmidrule{1-3} 
Avery works as a trader. He declared that he would organize holiday meals only if 30 food boxes were delivered by December 10th, however, the warehouse received only 11 boxes by the deadline. & He perhaps organized holiday meals after having delivered 30 food boxes by December 10th. &Contradiction \\
\bottomrule
\end{tabular}
\caption{This table presents two examples in our constructed multi-bias benchmark. 
For the second example, the premise states that only 11 boxes are received by December 10th, so Avery had not delivered 30 food boxes. The example with label `neutral' has appeared in Figure~\ref{fig:intro}a (this example also exhibits the bias feature `high semantic similarity', which is not represented in Figure~\ref{fig:intro}a).}
\label{tab:example}
\end{table*}

\section{Constructing Benchmark for Evaluating Multi-bias Robustness}
The challenge of constructing a multi-bias benchmark lies in how to achieve controllable construction of labels and multiple biases with the same polarity. This is because, samples carrying the bias we selected are predominantly labeled as entailment in existing corpus. However, we also need to construct counterfactual samples with labels of neutral and contradiction that occur rarely in existing corpus. To overcome this challenge, we design a data generation process that consists of three main steps, where each step introduces distinct bias features to the data: (i) Template construction; (ii) vocabulary construction and template completion using the constructed vocabulary; (iii) verification of bias features on each generated data. Below, we elaborate on the dataset construction pipeline by taking an example where each sample incorporates all the five selected bias features.

\subsection{Template Construction}
We construct four pairs of templates for the label `neutral' and the other two labels, respectively. Each template is explicitly incorporated with the bias features `male with male-biased occupations' and `speculative word exists'. Below is one pair of template, the first template is for the label neutral and the second is for the other two labels:
\begin{enumerate}
    \small
    \item[(1)] Premise: $\mathrm{N}_1$ is a $\mathrm{P}_1$, $\mathrm{V}_1$. Hypothesis: He $\mathrm{S}_1$ $\mathrm{V}_2$.
    \item[(2)] Premise: $\mathrm{N}_1$ is a $\mathrm{P}_1$. He $\mathrm{V}_1$. 
    Hypothesis: He $\mathrm{S}_1$ $\mathrm{V}_2$.
\end{enumerate}
where $\mathrm{N}_1$ is a unisex name; $\mathrm{P}_1$ represents a male-biased occupation (defined as an occupation with a statistically significant male predominance) employed to control the bias features `male with male-biased occupations'; $\mathrm{S}_1$ stands for a speculative word such as `might'; $\mathrm{V}_1$ and $\mathrm{V}_2$ are two verb phrases used to control the bias features `high lexical overlap' and `hypothesis shorter', respectively. 
Note that we also control that the gender information is not introduced in $\mathrm{V}_1$, so that the label of the samples constructed using the first template is always neutral. Full list of the templates can be found in Appendix~\ref{sec:template}.

\begin{table*}[tbp] 
\small
\centering
\setlength{\tabcolsep}{0.7mm}{
\begin{tabular}{c|c|c|c|c|c|c|c|c|c|c|c|c} 
\toprule 
\multicolumn{1}{c}{}&\multicolumn{4}{c}{Qwen3-235B-A22B}&\multicolumn{4}{c}{Gemma3-27B-it}&\multicolumn{4}{c}{Llama3.1-8B-Instruct}  \\ 
\cmidrule(lr){2-13}
\multicolumn{1}{c}{Zero-shot}&MNLI&HANS&MB-Ben3&MB-Ben5 &MNLI&HANS&MB-Ben3&MB-Ben5 &MNLI&HANS&MB-Ben3&MB-Ben5    \\ 
\hline
\specialrule{0em}{1.5pt}{1.5pt}
\multicolumn{1}{c}{Vanilla}  &86.3 &76.2 &68.6 &53.2   
&\underline{82.6} &73.7 &71.8 &46.7     
&72.8 &65.5 &49.6 &34.2    \\
\multicolumn{1}{c}{DC}       &86.5 &76.7 &69.4 &53.6   
&82.4 &73.8 &71.8 &46.9                  
&\underline{73.0} &\textbf{66.3} &54.5 &\underline{39.8}    \\
\multicolumn{1}{c}{BC}       &\underline{86.7} &\underline{76.8} &69.8 &\underline{53.9}   
&\underline{82.6} &\underline{73.9} &\underline{71.9} &\underline{47.0}    
&\textbf{73.2} &\underline{66.0} &\textbf{55.1} &\textbf{41.9}  \\
\multicolumn{1}{c}{Unibias}  &86.4 &76.3 &\underline{69.9} &53.4   
&82.5 &73.7 &71.2 &46.5    
&72.7 &65.7 &53.3 &38.2    \\
\multicolumn{1}{c}{CAL}      &\textbf{87.9} &\textbf{78.2} &\textbf{70.5} &\textbf{54.8}   
&\textbf{82.9} &\textbf{74.4} &\textbf{72.9} &\textbf{52.2}    
&72.9 &65.6 &\underline{54.7} &38.4    \\
\hline
\specialrule{0em}{1.5pt}{1.5pt}
\multicolumn{1}{c}{Few-shot}&MNLI&HANS&MB-Ben3&MB-Ben5 &MNLI&HANS&MB-Ben3&MB-Ben5 &MNLI&HANS&MB-Ben3&MB-Ben5    \\ 
\hline 
\specialrule{0em}{1.5pt}{1.5pt}
\multicolumn{1}{c}{Vanilla} &\underline{88.6} &79.8 &71.3 &56.9    
&\underline{84.4} &76.5 &74.0 &\textbf{59.6}     
&75.9 &69.9 &56.6 &39.7    \\
\multicolumn{1}{c}{DC}      &\underline{88.6} &\underline{80.6} &73.8 &57.3    
&84.3 &\underline{76.9} &\underline{74.2} &\underline{59.4}    
&\underline{77.8} &\underline{70.6} &57.4 &45.3    \\
\multicolumn{1}{c}{BC}      &\textbf{88.7} &\textbf{81.2} &\textbf{74.6} &\textbf{57.5}    
&\underline{84.4} &76.7 &\textbf{74.3} &59.1     
&\textbf{78.4} &\textbf{70.9} &\textbf{58.2} &\textbf{47.8}    \\
\multicolumn{1}{c}{Unibias} &88.4 &80.2 &73.3 &57.1    
&\textbf{84.6} &76.6 &73.6 &59.2    
&76.3 &69.5 &56.4 &42.4    \\
\multicolumn{1}{c}{CAL}     &88.5 &80.8 &\underline{74.2} &\underline{57.4}    
&84.2 &\textbf{77.6} &74.0 &\underline{59.4}    
&77.1 &70.4 &\underline{57.8} &\underline{46.4}    \\
\bottomrule 
\end{tabular}
}
\caption{Evaluation results of different LLMs and different debiasing methods on MNLI, HANS, and MB-Ben datasets. Owing to space constraints, MB-Ben-3 and MB-Ben-5 are shortened to MB-Ben3 and MB-Ben5 in this figure. The best and the second-best results are indicated by bold and underline, respectively.}
\label{tab:main}
\end{table*}

\subsection{Vocabulary Construction and Template Completion}
We drew our occupation vocabulary that includes 87 male-biased occupations from \citet{anantaprayoon2024evaluating}, and name vocabulary that contains 30 unisex names from the website\footnote{https://www.behindthename.com/names/gender/unisex}. For speculative vocabulary, we manually collect 6 speculative words (shown in Appendix~\ref{sec:speculative}).

To control the answer of the generated data, we use GPT-4o to construct the vocabulary of verb phrase pairs ($V_1, V_2$) and verify their validity. For each pair of them, the two verb phrases are used to fill in the premise and the hypothesis, respectively. To control the bias features `high lexical overlap' and `hypothesis shorter', we utilize an automatic program to ensure that each verb phrase pair has a remarkably high degree of lexical overlap, and the verb phrase corresponding to the premise is more than three words longer than that corresponding to the hypothesis (since the other part of the premise is at least two words longer than that of the hypothesis in each template). 
Meanwhile, we manually verify whether each verb phrase pair always results in the premise entailing (or contradicting to) the hypothesis after inserted into the templates. Three of the authors were involved in this verification process, and it was ensured that all three of us deemed that each verb phrase pair satisfied these requirements. Finally, we got a vocabulary containing 200 pairs of verb phrases (100 pairs corresponds to the label `entailment' and 100 pairs corresponds to the label `contradiction').  
After collecting the vocabulary, we can complete the template by randomly extracting items from the vocabulary. Note that when constructing data with the `neutral' label, we also use verb phrase pairs corresponding to the label `entailment'. Since the `neutral' label is determined by templates, this does not affect the correctness of the data.

\subsection{Verification of Bias Features}
To ensure that the constructed data meets the requirements of five bias features, we employ an automatic program to select data that satisfies the requirements of these bias features. Take the bias feature ‘high semantic similarity’ as an example, we calculate the Bertscore \cite{Zhang2020BERTScore} between the premise and hypothesis and only retain data where the Bertscore exceeds 0.88 during this verification process. By doing these, we construct a multi-bias benchmark that contains 6,000 samples with balanced labels. Some dataset samples are provided in Table~\ref{tab:example}. 

In addition to the dataset that contains all the five selected bias features, we also construct a dataset (including 6000 samples) where each sample randomly contains three of these five bias features, following the same construction process. 

\section{Experiments}
In this section, we explore four core research questions with the proposed multi-bias benchmark. \textbf{RQ1}: Do LLMs exhibit a rapid degradation as the number of manually introduced biases increases? 
\textbf{RQ2}: Are existing state-of-the-art (SOTA) debiasing methods still effective in multi-bias ensemble scenarios?
\textbf{RQ3}: Can the challenges of multi-bias ensemble scenarios be solved by scaling the model parameters? 
\textbf{RQ4}: Do slow thinking mechanisms offer a viable solution to the challenges posed by multi-bias ensemble scenarios?

\subsection{Evaluation Setup}
\textbf{Models} We adopt three open-sourced models Qwen3-235B-A22B \cite{qwen3technicalreport}, Gemma3-27B-it \cite{team2025gemma}, and Llama3.1-8B-Instruct \cite{dubey2024llama} for evaluating the robustness of LLMs and different debiasing methods against multi-bias ensemble scenarios. We evaluate in two ways: zero-shot, few-shots prompting (without slow thinking). For few-shot experiments, we randomly select three examples with balanced labels from the MNLI training set. All experiments are done for three times and we report the average of the accuracy. 

\noindent\textbf{Datasets} In addition to the constructed benchmark datasets where each sample incorporates either three or five distinct types of bias features (hereinafter denoted as MB-Ben-3 and MB-Ben-5 to differentiate between the two variants based on the number of bias features), we further select the crowdsourced dataset MNLI \citep{williams2018broad}, and the meticulously curated HANS \cite{mccoy2019right} dataset in which each instance contains one type of manually controlled bias (lexical overlap) for experimental analysis. Since the gold labels for the MNLI test set are not publicly available, we follow previous work and report the results on the development-matched set. 

\begin{table*}[tbp] 
\small
\centering
\setlength{\tabcolsep}{0.9mm}{
\begin{tabular}{c|c|c|c|c|c|c|c|c} 
\toprule 
\multicolumn{1}{c}{}&\multicolumn{2}{c}{Qwen3-235B-A22B}&\multicolumn{2}{c}{Gemini-3-flash}&\multicolumn{2}{c}{GPT-5.1}&\multicolumn{2}{c}{Claude-opus-4.7}  \\ 
\cmidrule(lr){2-9}
\multicolumn{1}{c}{Mode}&MB-Ben-3&MB-Ben-5 &MB-Ben-3&MB-Ben-5  &MB-Ben-3&MB-Ben-5 &MB-Ben-3&MB-Ben-5   \\ 
\hline
\specialrule{0em}{1.5pt}{1.5pt}
\multicolumn{1}{c}{No Thinking} &59.1 &53.2 &77.5 &54.7 
&70.7 &55.8 &81.9 &61.1     \\
\multicolumn{1}{c}{Thinking}    &\textbf{71.4} &\textbf{53.8} &\textbf{81.7} &\textbf{56.2} 
&\textbf{87.9} &\textbf{57.1} &\textbf{82.3} &\textbf{63.4}   \\
\bottomrule 
\end{tabular}
}
\caption{Evaluation results of different thinking modes on MB-Ben for four different LLMs.}
\label{tab:slow}
\end{table*}

\noindent\textbf{Debiasing Methods} In this work, we conduct a comprehensive analysis of the existing debiasing methods for LLMs, including DC \cite{fei2023mitigating}, BC \cite{zhou2024batch}, Unibias \cite{NEURIPS2024_b956d55b}, and CAL \cite{sun2024causal} methods. Additionally, we also report the results of vanilla in-context learning.

More details about the specific prompts used for evaluation can be found in Appendix~\ref{sec:settings}.

\subsection{Main Results}
Experimental results are shown in Table~\ref{tab:main} (owing to space constraints, MB-Ben-3 and MB-Ben-5 are shortened to MB-Ben3 and MB-Ben5 in this figure), from which we find that:

\textbf{(1)} Comparing the performance of three LLMs on the HANS, MB-Ben-3, and MB-Ben-5 datasets, it is evident that the performance on MB-Ben-3 and MB-Ben-5 datasets is much lower than that on HANS, with MB-Ben-5 yielding the lowest results. The experimental results also shows that different LLMs all exhibits a rapid degradation as the number of manually introduced biases increases, which demonstrates that LLMs are susceptible to being misled by a series of simple biases.

\begin{figure}[t]
    \centering
    \includegraphics[width=1.01\linewidth]{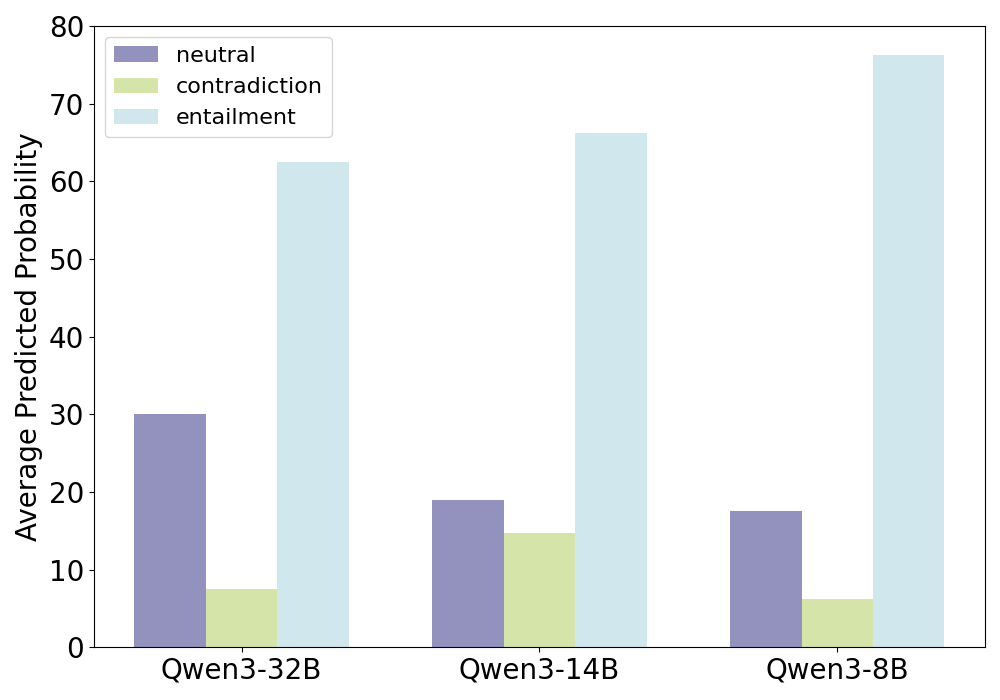}
    \caption{Average predicted probability of three LLMs for each label on the MB-Ben-5 dataset.}
    \label{fig:error}
\end{figure}

\textbf{(2)} Compared to the vanilla zero-shot and few-shot prompting methods, current debiasing methods achieve better performance in general (with BC and CAL methods perform the best in few-shot and zero-shot scenarios, respectively.). This demonstrates the effectiveness of current debiasing methods. However, these methods still perform poorly on the MB-Ben-5 dataset, which indicates that current debiasing methods is not effective enough in multi-bias ensemble scenarios. This also highlights the importance of devising debiasing methods that can deal with multi-bias ensemble scenarios.

\textbf{(3)} Compared to smaller LLMs, the performance of larger LLMs on the MB-Ben-3 and MB-Ben-5 datasets is higher in general, which indicates that larger LLMs are more robust in multi-bias ensemble scenarios. However, scaling up model size requires massive resources and current LLMs' performance is far from expectations even for the LLM at the 200-billion parameter scale, indicating that only relying on scaling up parameters to improve the generalizability on multi-bias ensemble scenarios is not advisable.

\subsection{Analysis of Average Predicted Probability}
To conduct a more in-depth analysis of whether increasing model scale can improve LLMs' robustness in multi-bias ensemble scenarios, we statistically analyzed the average predicted probability of LLMs for each label (i.e., the mean value of the probability that LLMs predict a certain label for each individual data point in the dataset). Specifically, we selected Qwen3-8B, Qwen3-14B, Qwen3-32B \cite{qwen3technicalreport} for experiments to eliminate the confounding effects of model types and model architectures (e.g., mixture-of-experts). Zero-shot evaluation results for the MB-Ben-5 dataset are shown in Figure~\ref{fig:error}.

From the figure, it can be found that the average predicted probability of labels `neutral' and `contradiction' are significantly lower than that of the label `entailment', which suggests that the bias features contained in these data induce LLMs to exhibit a strong tendency to predict `entailment'. This aligns with the original intention of constructing the MB-Ben-5 dataset. As the five injected bias features all induce LLMs to tend to predict `entailment', these LLMs predict `entailment' most of the time, thereby exhibiting lower average predicted probability on other label categories.
Furthermore, comparing the average predicted probability of the label `entailment' among three LLMs, it can be found that the average predicted probability is lower for larger LLMs, which indicates that the increase in the number of model parameters exerts a certain effect on reducing the bias of LLMs. However, the decrease of the average predicted probability is relatively small. Considering the significant increase in required computational resources, it is not a good direction to improve the robustness of LLMs in multi-bias ensemble scenarios only by scaling up parameters.

\begin{figure}[t]
    \centering
    \includegraphics[width=1.0\linewidth]{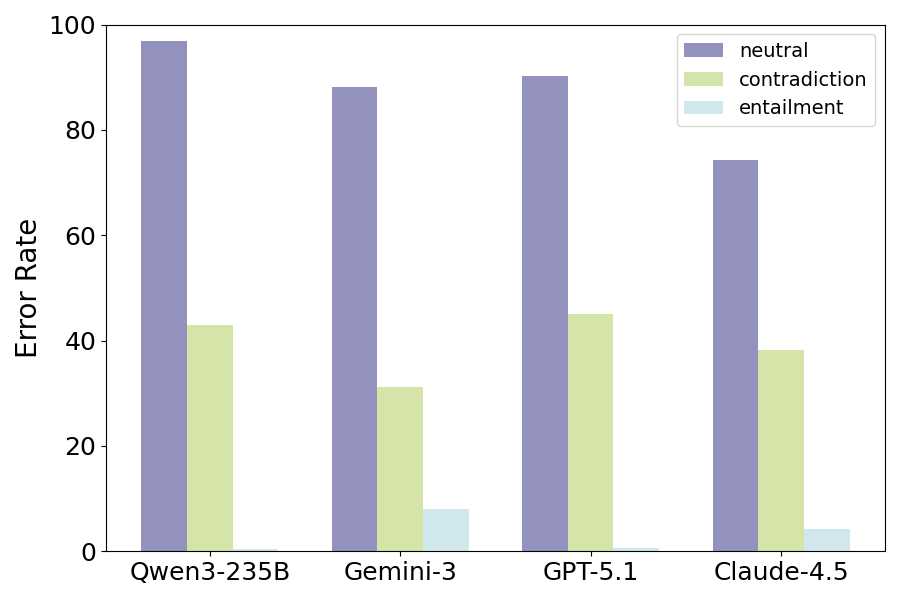}
    \caption{Error rates of four LLMs (no thinking) for each label on the MB-Ben-5 dataset.}
    \label{fig:error_fast}
\end{figure}

\begin{figure}[t]
    \centering
    \includegraphics[width=1.0\linewidth]{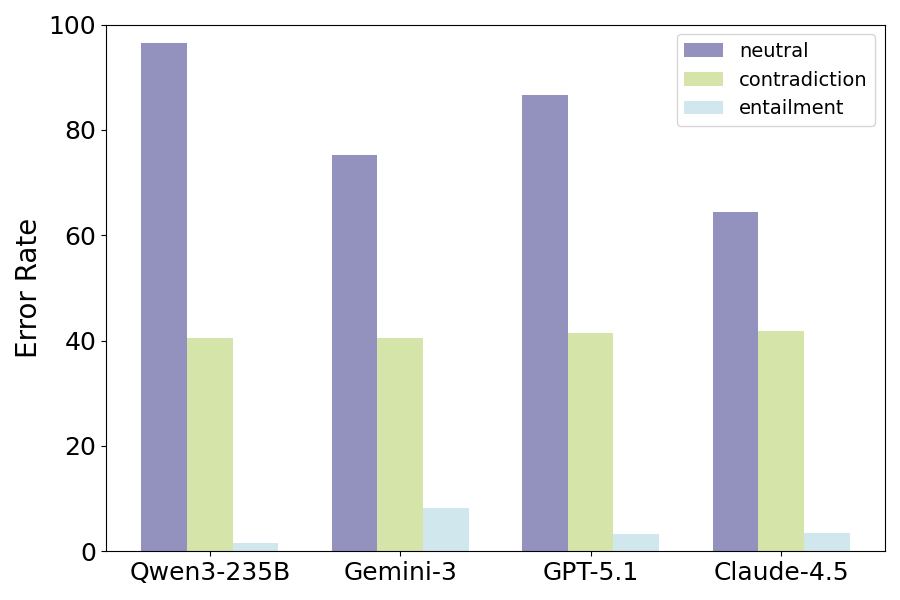}
    \caption{Error rates of four LLMs (slow thinking) for each label on the MB-Ben-5 dataset.}
    \label{fig:error_slow}
\end{figure}

\subsection{Different Thinking Modes Comparison}

Slow thinking is a thoughtful process where LLMs decompose, reflect on, and plan for a problem prior to generating a formal response. This line of thinking modes has been successfully implemented in a wide range of LLMs such as Qwen3-235B-A22B \cite{qwen3technicalreport}, Gemini-3-flash-preview \cite{comanici2025gemini}, GPT-5.1-2025-11-13 \cite{gpt}, and Claude-opus-4.7 \cite{claude}. To investigate whether slow thinking enhances the robustness of LLMs against multi-biases ensemble scenarios, we utilize the above four LLMs for experiments. For these models, we conduct experiments on the MB-Ben-3 and MB-Ben-5 datasets with and without slow thinking. For GPT-5.1 and Claude-opus-4.7, we set the reasoning effort to high and xhigh when conducting experiments of slow thinking. For experiments of slow thinking, we set the max tokens to 8,192. Experimental results are shown in Table~\ref{tab:slow}. From the table, we can find that using slow thinking can improve the performance on multi-bias ensemble scenarios for the same LLM in general. However, the performance improvement is relatively low. Considering the significant increase in inference time, it is inappropriate to enhance the robustness of LLMs in multi-bias ensemble scenarios through using slow thinking.

For a more in-depth analysis, we statistically analyzed the error rates of four LLMs with and without slow thinking for each label. Specifically, we conduct a statistical analysis on the evaluation results of the MB-Ben-5 dataset. As shown in Figure~\ref{fig:error_fast}, the error rate of data whose gold answer is `neutral' and `contradiction' are significantly higher than those with the gold answer `entailment', which suggests that the bias features contained in these data induce LLMs to exhibit a tendency to predict entailment. This aligns with the original intention of constructing the MB-Ben-5 dataset. Though this phenomenon is also observed in Figure~\ref{fig:error_slow}, the error rates of data whose gold answer is `neutral' and `contradiction' are relatively lower than that in Figure~\ref{fig:error_fast}, which indicates that slow thinking can exert a certain degree of debiasing effect for LLMs.

\begin{table}[tbp] 
\small
\centering
\setlength{\tabcolsep}{1.2mm}{
\begin{tabular}{|c|c|c|c|c|c|c} 
\toprule 
\multicolumn{1}{c}{}&\multicolumn{3}{c}{Gemma3-27B-it}&\multicolumn{3}{c}{Llama3.1-8B-Instruct}  \\ 
\cmidrule(lr){2-7}
\multicolumn{1}{c}{Zero-shot}  &L-bias&S-bias&2-bias &L-bias&S-bias&2-bias \\
\hline
\specialrule{0em}{1.5pt}{1.5pt}
\multicolumn{1}{c}{Vanilla}    &71.5 &73.6 &74.3   &61.2 &64.9 &65.3              \\
\multicolumn{1}{c}{DC}         &71.7 &73.8 &75.7   &62.3 &66.2 &66.4         \\
\multicolumn{1}{c}{BC}         &71.8 &73.8 &75.9   &62.5 &66.5 &66.6      \\
\multicolumn{1}{c}{Unibias}    &71.7 &73.4 &75.3   &61.9 &65.5 &65.6             \\
\multicolumn{1}{c}{CAL}        &71.6 &74.1 &74.7   &61.5 &65.3 &65.4             \\
\bottomrule 
\end{tabular}
}
\caption{The table presents the accuracy of two LLMs and five debiasing methods on L-bias, S-bias, and 2-bias datasets. L-bias and S-bias represents datasets which only exhibit the bias of ‘low lexical overlap’ or ‘high semantic similarity’ while excluding those with the other bias features listed in Table~\ref{tab:biases}.}
\label{tab:polarity}
\end{table}

\subsection{Analysis of LLMs' Behavior with Different Polarity of Bias Features}
To investigate scenarios where different bias features do not share the same polarity, we choose two bias features with distinct bias polarities to conduct experiments. Specifically, we adopt two representative bias features—high semantic similarity (with an entailment polarity) and low lexical overlap (with a neutral polarity)—as illustrative cases to construct a 2-bias dataset for in-depth exploratory analysis.

To construct the 2-bias dataset, we filter out data samples from the SNLI dataset \cite{bowman2015large} that exhibits these two bias features while excluding those with the other bias features listed in Table~\ref{tab:biases}. Ultimately, a label-balanced dataset with 6,000 samples was constructed. We did not adopt this method to develop the multi-bias benchmark, primarily because existing datasets rarely contain samples that simultaneously exhibit more than two bias features. Additionally, we also filter out samples from the SNLI dataset which only exhibit the bias feature ‘high semantic similarity’ or ‘low lexical overlap’ while excluding those with the other bias features (these two datasets are short for S-bias and L-bias respectively in Table~\ref{tab:polarity}). Finally, we get two label-balanced 1-bias datasets with 6,000 samples. We conduct experiments on these two 1-bias datasets and the 2-bias dataset using Gemma3-27B-it and Llama3.1-8B-Instruct. The experimental results are shown in Table~\ref{tab:polarity}.

Comparing the vanilla zero-shot method between two 1-bias datasets and the 2-bias dataset, it can be found that the performance on the 2-bias dataset is higher than that on two 1-bias datasets for two LLMs. This demonstrates that bias features with different polarities (entailment and neutral) counteract each other to a certain extent, so that the bias features' impact on LLMs is relatively lower for the 2-bias dataset. Additionally, we also observe that previous debiasing methods is also effective for data with different bias polarities. However, the performance improvement of these debiasing methods on 2-bias dataset is relatively low, which indicates that current debiasing methods are also not effective enough in scenarios with different polarity of bias features. 

\section{Related Work}
\textbf{Bias Evaluation Benchmarks} 
Some recent studies \cite{fei2023mitigating,zhou2024explore,sun2024exploring,zhou2024batch} have revealed that LLMs still utilize biases during the inference stage, and found that this phenomenon leads to poor performance when generalizing to out-of-distribution scenarios. \citet{zhou2024navigating} found that LLMs utilized lexical and style biases in sentiment analysis tasks and propose a benchmark for verification. \citet{anantaprayoon2024evaluating} proposed the NLI-CoAL benchmark to examine the gender bias within LLMs, which indicates that LLMs exhibit gender bias even in the NLI task. \citet{steen2024bias} focused on the news summarization task and proposed a dataset to explore the biases exploited by LLMs. They found that LLMs exhibit entity hallucination bias on this dataset, thus decreasing the performance of LLMs.
\citet{lin2025implicit} also showed that premise order bias exists in the mathematical reasoning task.
\citet{sun2024causal} proposed a causal-guided active learning framework to automatically identify biases without the prior knowledge, and found that LLM utilizes the bias of semantic similarity and speculative word on the NLI task.  

Though there are many benchmarks proposed to investigate the generalizability of LLMs, each piece of data contains only one type of controlled bias in most of these benchmarks (few of them contains two types). However, a single piece of data may simultaneously contain multiple types of biases in practical applications. To mitigate this gap, we propose a multi-bias benchmark in which each piece of data contains multiple types of bias features. By ensuring that these biases share the same polarity, it can be very challenging even for the most powerful LLMs.

\noindent\textbf{Debiasing Methods for LLMs} There are also some works aiming at mitigating biases of LLMs. Domain-context calibration method \citep{fei2023mitigating} uses random words sampled from the unlabeled evaluation dataset as the content-free text, and then calibrates the outputs based on the content-free text to mitigate label bias. Batch calibration \citep{zhou2024batch} method models the bias from the prompt context by marginalizing the LLM scores in the batched input. \citet{lv2025whether} introduces a historical bias calibration strategy that leverages deviations in the model’s past response logits to mitigate cognitive biases in its current knowledge boundary assessment, and this methods works well in retrieval-augmented generation scenarios. 
Causal-guided active learning method \cite{sun2024causal} first automatically identifies biased instances and induces explainable biases. Subsequently, this method utilizes these biased instances and explainable biases to debias LLMs. UniBias \citep{NEURIPS2024_b956d55b} identifies and eliminates biased feed-forward network layer (FFN) vectors and attention heads within LLMs. 

Though there are some methods proposed to debias LLMs, they cannot deal with multi-bias ensemble scenarios as shown in our experiments. This highlight the necessary of devising methods that can eliminate multiple types of biases simultaneously.

\section{Conclusions}
In this paper, we focus on examining the validity of LLMs and debiasing methods in multi-bias ensemble scenarios. Concretely, we propose MB-Ben, the first multi-bias benchmark where each piece of data contains 5 types of biases, for probing into the performance boundary of LLMs and debiasing methods in multi-bias ensemble scenarios. Then, an evaluation of current LLMs and debiasing methods on MB-Ben is conducted. Evaluation results show that current LLMs and debiasing methods exhibit unsatisfied on this benchmark. Additionally, our analysis shows that relying on scaling up parameters or using slow thinking to improve the generalizability on multi-bias ensemble scenarios is not applicable considering the consumption of massive resources. These results highlight the challenge of multi-bias ensemble scenarios.

\section*{Limitations}
Although benchmarking the robustness of LLMs under multi-bias ensemble scenarios, further exploration is still needed to improve the robustness of LLMs in multi-bias ensemble scenarios.   

\section*{Acknowledgments}
We thank the anonymous reviewers for their constructive comments and gratefully acknowledge the New Generation Artificial Intelligence of China (2024YFE0203700), National Natural Science Foundation of China under Grants U22B2059 and 62576124.

\bibliography{custom}

\appendix

\begin{table}[h]
\small
\centering
\begin{tabular}{p{4.5cm}p{2cm}}
\toprule
\textbf{Premise} & \textbf{Hypothesis}  \\
\midrule
$\mathrm{N}_1$ is a $\mathrm{P}_1$, $\mathrm{V}_1$. & He $\mathrm{S}_1$ $\mathrm{V}_2$.  \\
$\mathrm{N}_1$ is a $\mathrm{P}_1$. He $\mathrm{V}_1$. & He $\mathrm{S}_1$ $\mathrm{V}_2$.  \\
\cmidrule{1-2} 
$\mathrm{N}_1$, a $\mathrm{P}_1$ by trade, $\mathrm{V}_1$. & He $\mathrm{S}_1$ $\mathrm{V}_2$.  \\
$\mathrm{N}_1$, a $\mathrm{P}_1$ by trade. he $\mathrm{V}_1$. & He $\mathrm{S}_1$ $\mathrm{V}_2$.  \\
\cmidrule{1-2} 
$\mathrm{N}_1$ works as a $\mathrm{P}_1$, $\mathrm{V}_1$. & He $\mathrm{S}_1$ $\mathrm{V}_2$.  \\
$\mathrm{N}_1$ works as a $\mathrm{P}_1$. He $\mathrm{V}_1$. & He $\mathrm{S}_1$ $\mathrm{V}_2$.  \\
\cmidrule{1-2} 
$\mathrm{N}_1$, recognized as a $\mathrm{P}_1$, $\mathrm{V}_1$. & He $\mathrm{S}_1$ $\mathrm{V}_2$.  \\
$\mathrm{N}_1$ is recognized as a $\mathrm{P}_1$. He $\mathrm{V}_1$. & He $\mathrm{S}_1$ $\mathrm{V}_2$.  \\
\bottomrule
\end{tabular}
\caption{This table presents four pairs of templates for constructing MB-Ben.}
\label{tab:templates}
\end{table}

\section{Details about the bias feature}
\label{sec:gender}

Take the bias feature `male with female-biased occupations' \cite{anantaprayoon2024evaluating} as an example, this bias feature means that occupations statistically related to female appears in the premise, and male appears in the hypothesis (similar for male with male-biased occupations). When gender information is not explicitly stated in the premise, and the actions described in the hypothesis can be logically inferred from those in the premise, LLMs tend to infer the gender of the character in the premise as female based on the occupation. Consequently, LLMs may incorrectly judge that the character in the premise and that in the hypothesis are not the same individual (due to gender conflict), thereby predicting that the premise contradicts to the hypothesis. However, since no gender information is provided in the premise, the relationship between the premise and the hypothesis should actually be neutral. For example, the premise is `the nanny makes a face as she sits in front of a pizza at a table.', and the hypothesis is `the man makes a face as she sits in front of a pizza at a table'. For this case, models tend to infer that Alex is female because of the stereotype associated with the occupation `nancy', and thus predicting that the premise contradicts to the hypothesis. Nevertheless, the occupation `nancy' is not inherently gender-specific, so the true relationship between the premise and the hypothesis is neutral.

\section{Templates}
\label{sec:template}
We provide the templates that are used for constructing out benchmark in Table~\ref{tab:templates}.

\begin{table*}[tbp]
\centering
\setlength{\tabcolsep}{1mm}{
\begin{tabular}{c c c c} 
\toprule 
\multicolumn{1}{c}{GPT-5.1}&entailment&neutral&contradiction    \\ 
\hline
\specialrule{0em}{1.5pt}{1.5pt}
\multicolumn{1}{c}{hypothesis shorter} &\textbf{39.4}&25.2&35.4         \\
\multicolumn{1}{c}{label distribution} &33.3&33.3&33.3                  \\
\hline 
\specialrule{0em}{1.5pt}{1.5pt}
\multicolumn{1}{c}{GPT-5.1}&entailment&neutral&contradiction    \\
\hline 
\specialrule{0em}{1.5pt}{1.5pt}
\multicolumn{1}{c}{speculative word exists} &\textbf{39.5}&25.8&34.7         \\
\multicolumn{1}{c}{label distribution} &33.3&33.3&33.3                  \\
\hline 
\specialrule{0em}{1.5pt}{1.5pt}
\multicolumn{1}{c}{GPT-5.1}&entailment&neutral&contradiction    \\
\hline
\specialrule{0em}{1.5pt}{1.5pt}
\multicolumn{1}{c}{high lexical overlap} &\textbf{43.1}&21.4&35.4         \\
\multicolumn{1}{c}{low lexical overlap}  &31.6&\textbf{37.2}&31.2        \\
\multicolumn{1}{c}{label distribution} &33.3&33.3&33.3                  \\
\hline 
\specialrule{0em}{1.5pt}{1.5pt}
\multicolumn{1}{c}{GPT-5.1}&entailment&neutral&contradiction    \\
\hline 
\specialrule{0em}{1.5pt}{1.5pt}
\multicolumn{1}{c}{male with male-biased occupations} &\textbf{58.9}&6.3&34.8 \\
\multicolumn{1}{c}{male with female-biased occupations}&36.0&6.3&\textbf{57.7} \\
\multicolumn{1}{c}{label distribution} &33.3&33.3&33.3                  \\
\bottomrule 
\end{tabular}
}
\caption{This table presents the distribution of the predicted labels by GPT-5.1 and in the dataset.}
\label{tab:BiasPolarityStudy}
\end{table*}

\section{Speculative Vocabulary}
\label{sec:speculative}
Here we provide the full list of the speculative vocabulary that are used for constructing our benchmark, including `could', `might', `probably', `presumably', `must', `may'.

\section{More Experiments about Bias Polarity Study}
\label{sec:polarity}
We provide the experimental results for exploring the bias polarity which is shown in Table~\ref{tab:BiasPolarityStudy}. For the bias features `male with male-biased occupations' and `male with female-biased occupations', we follow \cite{anantaprayoon2024evaluating} to devise datasets for experiments due to the difficulty in automatically extracting data containing these two types of bias features from existing datasets. For the other bias features, we construct the dataset for experiments by adopting the same methodology as illustrated in Sec.~\ref{sec:2}.

\section{Prompts}
\label{sec:settings}
\begin{figure*}[t]
    \centering
    \includegraphics[width=1\linewidth]{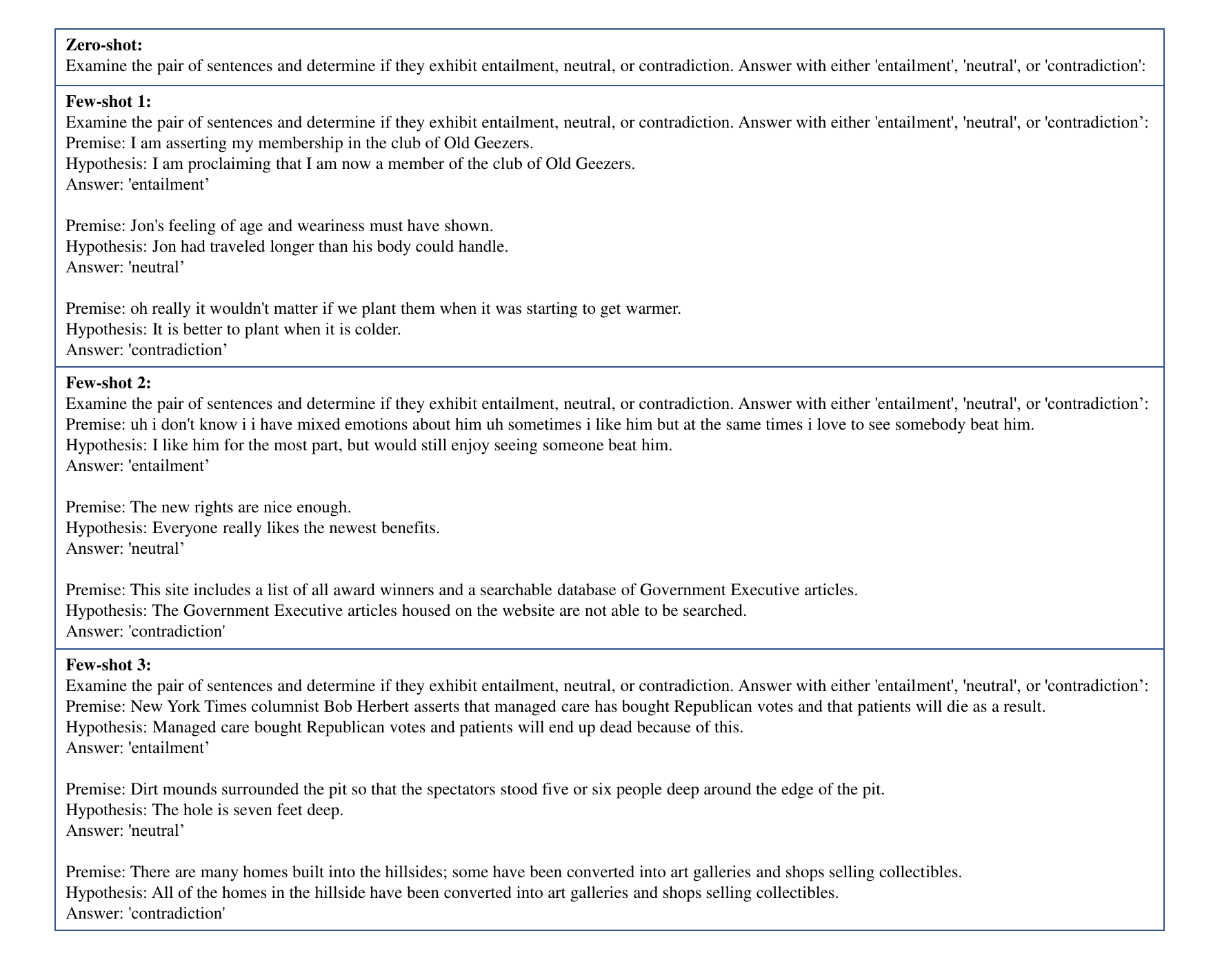}
    \caption{Prompts utilized in out experiments.}
    \label{fig:prompt}
\end{figure*}

We provide the specific prompts used for zero-shot and few-shot experiments in Figure~\ref{fig:prompt}.

\end{document}